%% file: emnlp2016.tex
\documentclass[11pt,letterpaper]{article}
\usepackage{emnlp2016}
\usepackage{times}
\usepackage{latexsym}
\usepackage{graphicx}
\usepackage[caption=false]{subfig}
\usepackage{float}
\usepackage{color}
\usepackage{lipsum}
\usepackage{enumitem}
\setlist{nolistsep}
\usepackage{arydshln}

\def\figref#1{Fig.~\ref{#1}}
\def\secref#1{Sec.~\ref{#1}}
\def\tabref#1{Table~\ref{#1}}

\emnlpfinalcopy

\def\emnlppaperid{993}

\title{Interpreting Neural Networks to Improve Politeness Comprehension}

\author{
Malika Aubakirova \\ University of Chicago \\ {\tt aubakirova@uchicago.edu}
\And  
Mohit Bansal \\ UNC Chapel Hill \\ {\tt mbansal@cs.unc.edu}
}

\date{}

\begin{document}

\maketitle
\input{abstract}
\input{introduction}
\input{relatedwork}
\input{approach}
\input{tasksetup}

\input{results}
\input{analysisvisualizations}
\input{conclusion}
\input{acknowledgment}
\appendix
\vspace{15pt}
\noindent
{\fontsize{12}{12}\selectfont \textbf{Appendix}} \par
\vspace{5pt}
\input{supplementary}

\bibliography{politenessrefs}
\bibliographystyle{emnlp2016}

\end{document}

%% file: abstract.tex
\begin{abstract}
We present an interpretable neural network approach to predicting and understanding politeness in natural language requests. Our models are based on simple convolutional neural networks directly on raw text, avoiding any manual identification of complex sentiment or syntactic features, while performing better than such feature-based models from previous work. More importantly, we use the challenging task of politeness prediction as a testbed to next present a much-needed understanding of what these successful networks are actually learning. For this, we present several network visualizations based on activation clusters, first derivative saliency, and embedding space transformations, helping us automatically identify several subtle linguistics markers of politeness theories.
Further, this analysis reveals multiple novel, high-scoring politeness strategies which, when added back as new features, reduce the accuracy gap between the original featurized system and the neural model, thus providing a clear quantitative interpretation of the success of these neural networks.

\end{abstract}

%% file: introduction.tex
\section{Introduction}

Politeness theories~\cite{brown1987politeness,gu1990politeness,bargiela2003face} include key components such as modality, indirection, deference, and impersonalization.
Positive politeness strategies focus on making the hearer feel good through offers, promises, and jokes. Negative politeness examples include favor seeking, orders, and requests. 
Differentiating among politeness types is a highly nontrivial task, because it depends on factors such as a context, relative power, and culture. \newcite{danescu2013computational} proposed a useful computational framework for predicting politeness in natural language requests by designing various lexical and syntactic features about
key politeness theories, e.g., first or second person start vs. plural. 
However, manually identifying such politeness features is very challenging, because there exist several complex theories and politeness in natural language is often realized via subtle markers and non-literal cues. 

Neural networks have been achieving high performance in sentiment analysis tasks, via their ability to automatically learn short and long range spatial relations. However, it is hard to interpret and explain what they have learned.
In this paper, we first propose to address politeness prediction via simple CNNs working directly on the raw text. This helps us avoid the need for any complex, manually-defined linguistic features, while still performing better than such featurized systems. More importantly, we next present an intuitive interpretation of what these successful neural networks are learning, using the challenging politeness task as a testbed. 

To this end, we present several visualization strategies: activation clustering, first derivative saliency, and embedding space transformations, some of which are inspired by similar strategies in computer vision~\cite{erhan2009visualizing,simonyan2013deep,girshick2014rich}, and have also been recently adopted in NLP for recurrent neural networks~\cite{li2015visualizing,kadar2016rep}. 
The neuron activation clustering method not only rediscovers and extends several manually defined features from politeness theories, but also uncovers multiple novel strategies, whose importance we measure quantitatively. 
The first derivative saliency technique allows us to identify the impact of each phrase on the final politeness prediction score via heatmaps, revealing useful politeness markers and cues. 
Finally, we also plot lexical embeddings before and after training, showing how specific politeness markers move and cluster based on their polarity. Such visualization strategies should also be useful for understanding similar state-of-the-art neural network models on various other NLP tasks.

Importantly, our activation clusters reveal two novel politeness strategies, namely indefinite pronouns and punctuation. Both strategies display high politeness and top-quartile scores (as defined by~\newcite{danescu2013computational}). Also, when added back as new features to the original featurized system, they improve its performance and reduce the accuracy gap between the featurized system and the neural model, thus providing a clear, quantitative interpretation of the success of these neural networks in automatically learning useful features.

%% file: relatedwork.tex
\section{Related Work}
\newcite{danescu2013computational} presented one of the first useful datasets and computational approaches to politeness theories~\cite{brown1987politeness,goldsmith2007brown,kadar2013understanding,locher2005politeness}, using manually defined lexical and syntactic features.
Substantial previous work has employed machine learning models for other sentiment analysis style  tasks~\cite{pang2002thumbs,pang2004sentimental,kennedy2006sentiment,go2009twitter,ghiassi2013twitter}. 
Recent work has also applied neural network based models to sentiment analysis tasks~\cite{chen2011neural,socher2013recursive,moraes2013document,dong2014adaptive,dos2014deep,kalchbrenner2014convolutional}. 
However, none of the above methods focused on visualizing and understanding the inner workings of these successful neural networks.

There have been a number of visualization techniques explored for neural networks in computer vision~\cite{krizhevsky2012imagenet,simonyan2013deep,zeiler2014visualizing,samek2015evaluating,mahendran2015understanding}.
Recently in NLP,~\newcite{li2015visualizing} successfully adopt computer vision techniques, namely first-order \textit{saliency}, and present representation plotting for sentiment compositionality across RNN variants. Similarly,~\newcite{kadar2016rep} analyze the omission scores and top-k contexts of hidden units of a multimodal RNN. \newcite{karpathy2015visualizing} visualize character-level language models. We instead adopt visualization techniques for CNN style models for NLP\footnote{The same techniques can also be applied to RNN models.} and apply these to the challenging task of politeness prediction, which often involves identifying subtle and non-literal sociolinguistic cues. We also present a quantitative interpretation of the success of these CNNs on the politeness prediction task, based on closing the performance gap between the featurized and neural models.

%% file: approach.tex
\section{Approach}
 
\subsection{Convolutional Neural Networks}
We use one convolutional layer followed by a pooling layer. For a sentence $ v_{1:n} $ (where each word $v_i$ is a $d$-dim vector), a filter $ m $ applied on a window of $ t $ words, produces a convolution feature $ c_i = f(m * v_{i:i+t-1}+b) $,
where $ f $ is a non-linear function, and $ b $ is a bias term. A \textit{feature map} $ c \in R^{n-t+1}$ is applied on each possible window of words so that $ c = [ c_{1}, ..., c_{n-t+1} ] $. This convolutional layer is then followed by a max-over-pooling operation~\cite{collobert2011natural} that gives $ C = max \{c\} $ of the particular filter.   
To obtain multiple features, we use multiple filters of varying window sizes. The result is then passed to a fully-connected softmax layer that outputs probabilities over labels.

%% file: tasksetup.tex
\section{Experimental Setup}

\subsection{Datasets}
We used the two datasets released by \newcite{danescu2013computational}: Wikipedia (Wiki) and Stack Exchange (SE), containing community requests with politeness labels.
Their `feature development' was done on the Wiki dataset, and SE was used as the `feature transfer' domain. 
We use a simpler train-validation-test split based setup for these datasets instead of the original leave-one-out cross-validation setup, which makes training extremely slow for any neural network or sizable classifier.\footnote{The result trends and visualizations using cross-validation were similar to our current results, in preliminary experiments. We will release our exact dataset split details.}

\subsection{Training Details}
Our tuned hyperparameters values (on the dev set of Wiki) are a mini-batch size of 32, a learning rate of 0.001 for the Adam~\cite{kingma-15} optimizer, a dropout rate of 0.5, CNN filter windows of 3, 4, and 5 with 75 feature maps each, and ReLU as the non-linear function~\cite{nair2010rectified}. For convolution layers, we use valid padding and strides of all ones. We followed~\newcite{danescu2013computational} in using SE only as a ‘transfer’ domain, i.e., we do not re-tune any hyperparameters or features on this domain and simply use the chosen values from the Wiki setting.
The split and other training details are provided in the supplement.

\begin{figure*}[t]
\vspace{-5pt}
\begin{center}
\setlength{\tabcolsep}{5pt}
\renewcommand{\arraystretch}{0}
\begin{tabular}{c c c }
\includegraphics[width=5cm]{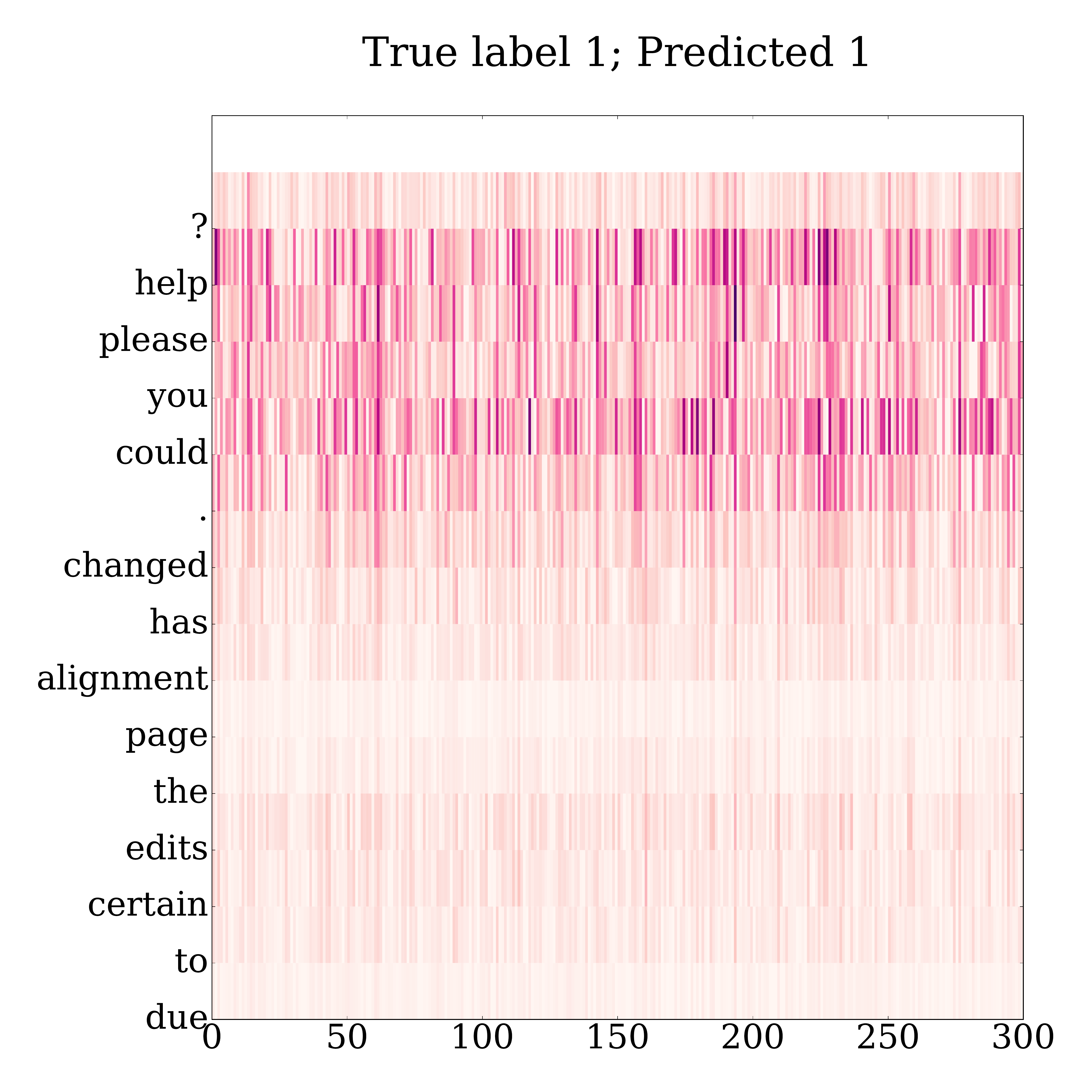} & \includegraphics[width=5cm]{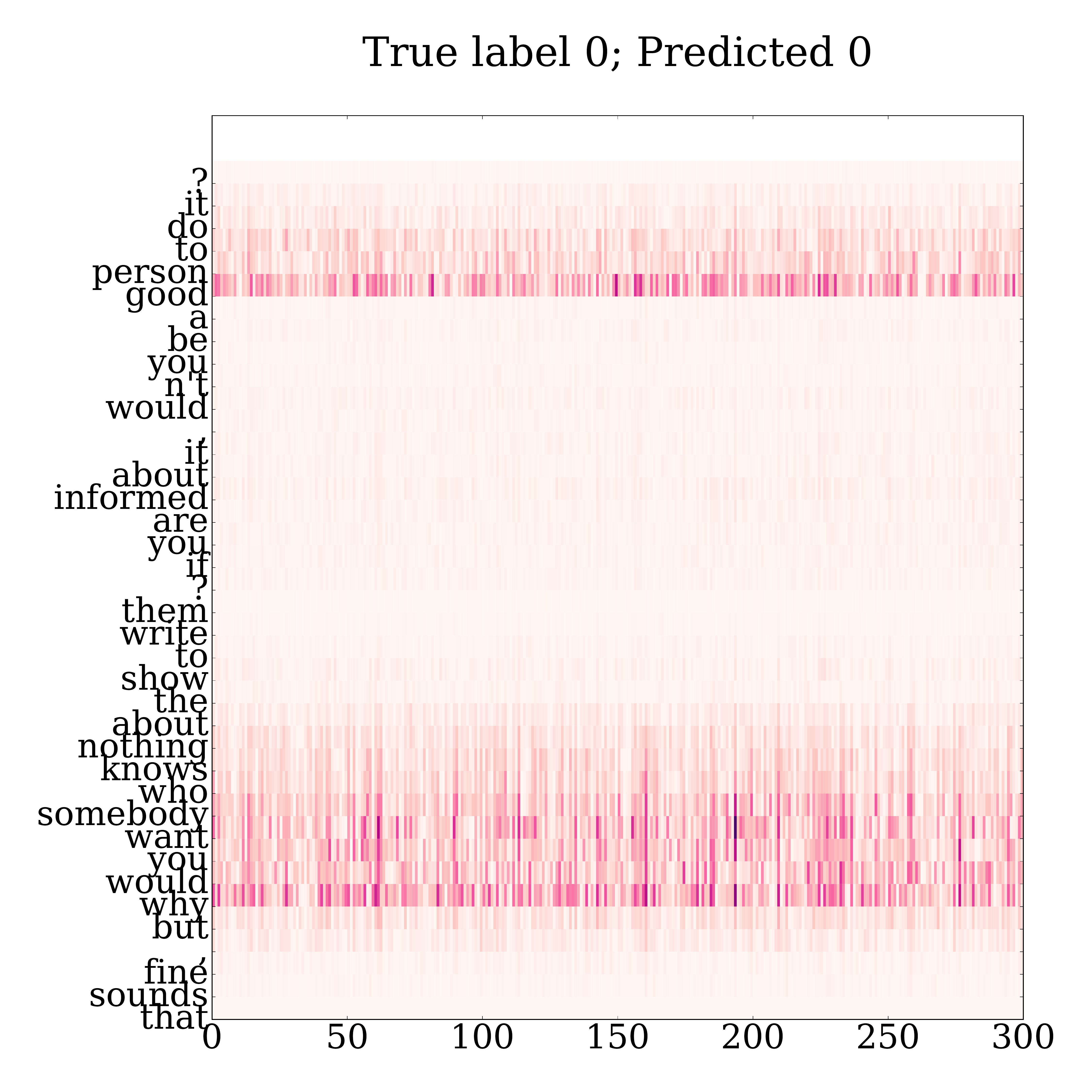} &  \includegraphics[width=5cm]{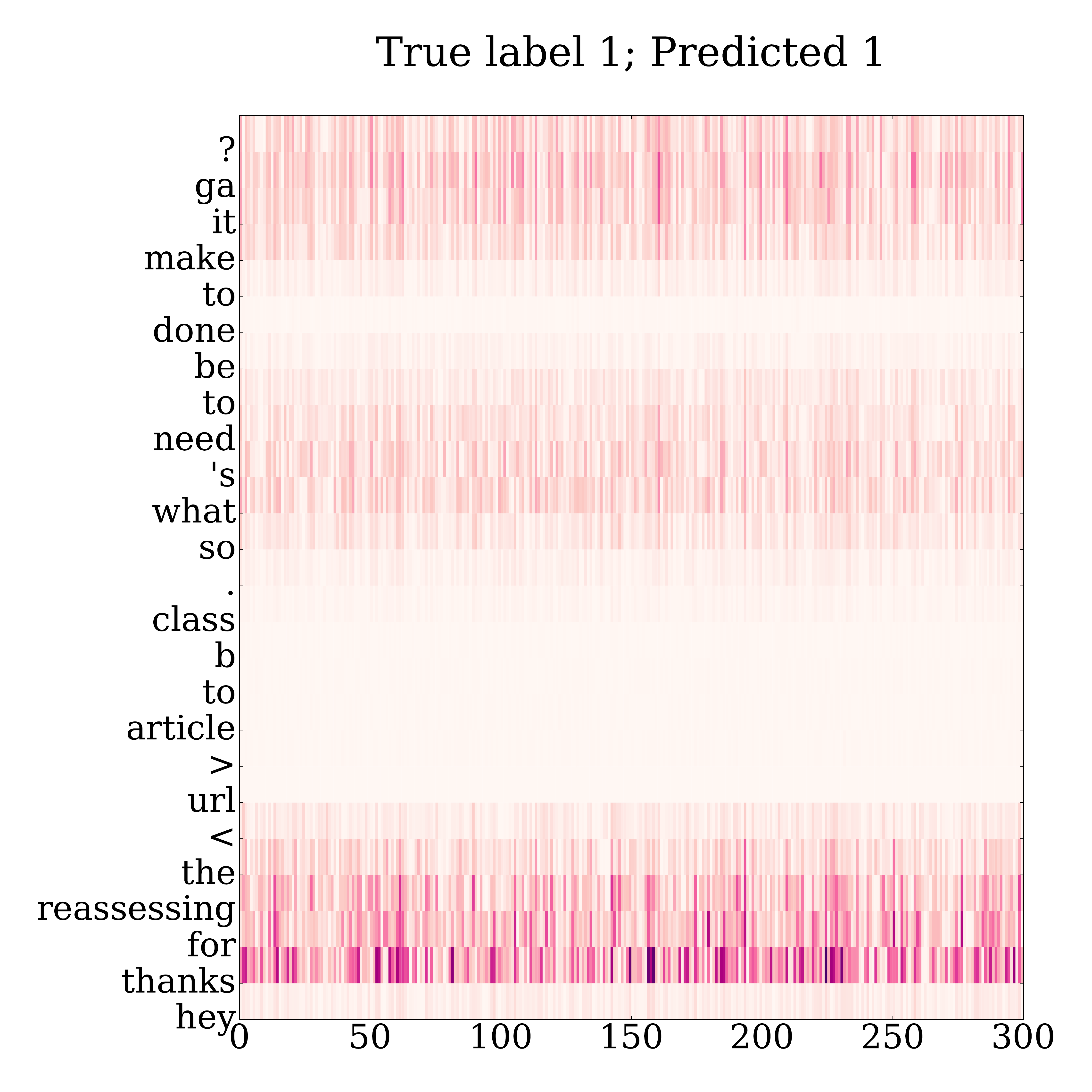}\\
\end{tabular}
\end{center}
\vspace{-7pt}
\caption{\label{tab:heatmap_table}Saliency heatmaps for correctly classified sentences.}
\vspace{-10pt}
\end{figure*}

%% file: results.tex
\section{Results}

\tabref{tab:results_table} first presents our reproduced classification accuracy test results (two labels: positive or negative politeness) for the  bag-of-words and linguistic features based models of~\newcite{danescu2013computational} (for our dataset splits) as well as the performance of our CNN model. As seen, without using any manually defined, theory-inspired linguistic features, the simple CNN model performs better than the feature-based methods.\footnote{For reference, human performance on the original task setup of~\newcite{danescu2013computational} was 86.72\% and 80.89\% on the Wiki and SE datasets, respectively.}

Next, we also show how the linguistic features baseline improves on adding our novelly discovered features (plus correcting some exising features), revealed via the analysis in \secref{sec:analysis}. Thus, this reduces the gap in performance between the linguistic features baseline and the CNN, and in turn provides a quantitative reasoning for the success of the CNN model. More details in~\secref{sec:analysis}.

%% file: analysisvisualizations.tex
\section{Analysis and Visualization}
\label{sec:analysis}
We present the primary interest and contribution of this work: performing an important qualitative and quantitative analysis of what is being learned by our neural networks w.r.t. politeness strategies.\footnote{We only use the Wiki train/dev sets for all analysis.}

\begin{table}
\begin{center}
\begin{small}
\begin{tabular}{| l | c | c | }
\hline
\textbf{Model} & \textbf{Wiki} & \textbf{SE} \\ \hline
\hline
Bag-of-Words & \ \ 80.9\% \ \ & \ \ 64.6\% \ \ \\
Linguistic Features  \ \ \   & 82.6\% & 65.2\% \\
With Discovered Features & 83.8\% & 65.7\% \\
CNN & 85.8\% & 66.4\% \\
\hline
\end{tabular}
\end{small}
\end{center}
\vspace{-3pt}
\caption{\label{tab:results_table}Accuracy Results on Wikipedia and Stack Exchange.}
\vspace{-12pt}
\end{table}

\subsection{Activation Clusters}
\label{clusterssection}
Activation clustering is a non-parametric approach (adopted from~\newcite{girshick2014rich}) of computing each CNN unit's activations on a dataset and then analyzing the top-scoring samples in each cluster.
We keep track of which neurons get maximally activated for which Wikipedia requests and analyze the most frequent requests in each neuron's cluster, to understand what each neuron reacts to.

\subsubsection{Rediscovering Existing Strategies}
We find that the different activation clusters of our neural network automatically rediscover a number of strategies from politeness theories considered in~\newcite{danescu2013computational} (see Table 3 in their paper). We present a few such strategies here with their supporting examples, and the rest (e.g., Gratitude, Greeting, Positive Lexicon, and Counterfactual Modal) are presented in the supplement. The majority politeness label of each category is indicated by (+) and (-).

\noindent\textbf{Deference (+)} $\;$ A way of sharing the burden of a request placed on the addressee.
Activation cluster examples: \{``\textit{nice work so far on your rewrite...}"; ``\textit{hey, good work on the new pages...}"\}

\noindent\textbf{Direct Question (-)} $\;\;$ Questions imposed on the converser in a direct manner with a demand of a factual answer. 
Activation cluster examples: \{``\textit{what's with the radio , and fist in the air?}"; ``\textit{what level warning is appropriate?}"\}

\subsubsection{Extending Existing Strategies}
We also found that certain activation clusters depicted interesting extensions of the politeness strategies given in previous work. 

\noindent\textbf{Gratitude (+)} $\;\;$ Our CNN learns a special shade of gratitude, namely it distinguishes a cluster consisting of the bigram \textit{thanks for}.
Activation cluster examples: \{``\textit{thanks for the good advice.}"; ``\textit{thanks for letting me know.}"\}

\noindent\textbf{Counterfactual Modal (+)} $\;\;$ Sentences with \textit{Would you/Could you} get grouped together as expected; but in addition, the cluster contains requests with \textit{Do you mind} as well as gapped 3-grams like \textit{Can you ... please?}, which presumably implies that the combination of a later \textit{please} with future-oriented variants \textit{can/will} in the request gives a similar effect as the conditional-oriented variants \textit{would/could}. 
Activation cluster examples: \{\textit{can this be reported ... grid, please?}"; \textit{do you mind having another look?}"\} 

\subsubsection{Discovering Novel Strategies}
In addition to rediscovering and extending politeness strategies mentioned in previous work, our network also automatically discovers some novel activation clusters, potentially corresponding to new politeness strategies. 

\noindent\textbf{Indefinite Pronouns (-)} $\;\;$ \newcite{danescu2013computational} distinguishes requests with first and second person (plural, starting position, etc.).
However, we find activations that also react to indefinite pronouns such as \textit{something/somebody}.
Activation cluster examples: \{``\textit{am i missing something here?}"; ``\textit{wait for anyone to discuss it.}"\} 

\noindent\textbf{Punctuation (-)} $\;\;$ Though non-characteristic in direct speech, punctuation appears to be an important special marker in online communities, which in some sense captures verbal emotion in text. 
E.g., one of our neuron clusters gets activated on question marks ``???" and one on ellipsis ``...".
Activation cluster examples:  \{``\textit{now???}"; ``\textit{original article????}"; ``\textit{helllo?????}"\}\footnote{More examples are given in the supplement.}

In the next section, via saliency heatmaps, we will further study the impact of indefinite pronouns in the final-decision making of the classifier. Finally, in \secref{sec:quant}, we will quantitatively show how our newly discovered strategies help directly improve the accuracy performance of the linguistic features baseline and achieve high politeness and top-quartile scores as per~\newcite{danescu2013computational}.

\subsection{First Derivative Saliency}
\label{FDS}
Inspired from neural network visualization in computer vision ~\cite{simonyan2013deep}, the first derivative saliency method indicates how much each input unit contributes to the final decision of the classifier. If $E$ is the input embedding, $y$ is the true label, and $S_y(E)$ is the neural network output, then we consider gradients $\frac{\partial S_y(E)}{\partial e} $. Each image in \figref{tab:heatmap_table} is a heatmap of the magnitudes of the derivative in absolute value with respect to each dimension. 

The first heatmap gets signals from \textit{please} (Please strategy) and \textit{could you} (Counterfactual Modal strategy), but effectively puts much more mass on \textit{help}. This is presumably due to the nature of Wikipedia requests such that the meaning boils down to asking for some help that reduces the social distance. 
In the second figure, the highest emphasis is put on \textit{why would you}, conceivably used by Wikipedia administrators as an indicator of questioning.
Also, the indefinite pronoun \textit{somebody} makes a relatively high impact on the decision.
This relates back to the activation clustering mentioned in the previous section, where indefinite pronouns had their own cluster.
In the third heatmap, the neural network does not put much weight on the greeting-based start \textit{hey}, because it instead focuses on the higher polarity\footnote{See Table 3 of ~\newcite{danescu2013computational} for polarity scores of the various strategies.} gratitude part after the greeting, i.e., on the words \textit{thanks for}. This will be further connected in \secref{EST}.

\begin{table*}[t]
\begin{center}
\begin{small}
\begin{tabular}{| l | c | c | c | c| c|}
\hline
& \textbf{Strategy} & \textbf{Politeness} & \textbf{In top quartile} & \textbf{Examples} \\ \hline
21. & Indefinite Pronouns & -0.13 & 39\% & \textit{am i missing something here?}  \\ \hline
22. & Punctuation & -0.71 & 62\% & \textit{helllo?????} \\
\hline
\end{tabular}
\end{small}
\end{center}
\caption{\label{tab:extend_table} Extending Table 3 of Danescu-Niculescu-Mizil et al. (2013) with our novelly discovered politeness strategies.}
\vspace{-12pt}
\end{table*}

\subsection{Embedding Space Transformations}
\label{EST}

\begin{figure}[t]
\includegraphics[width=7.9cm]{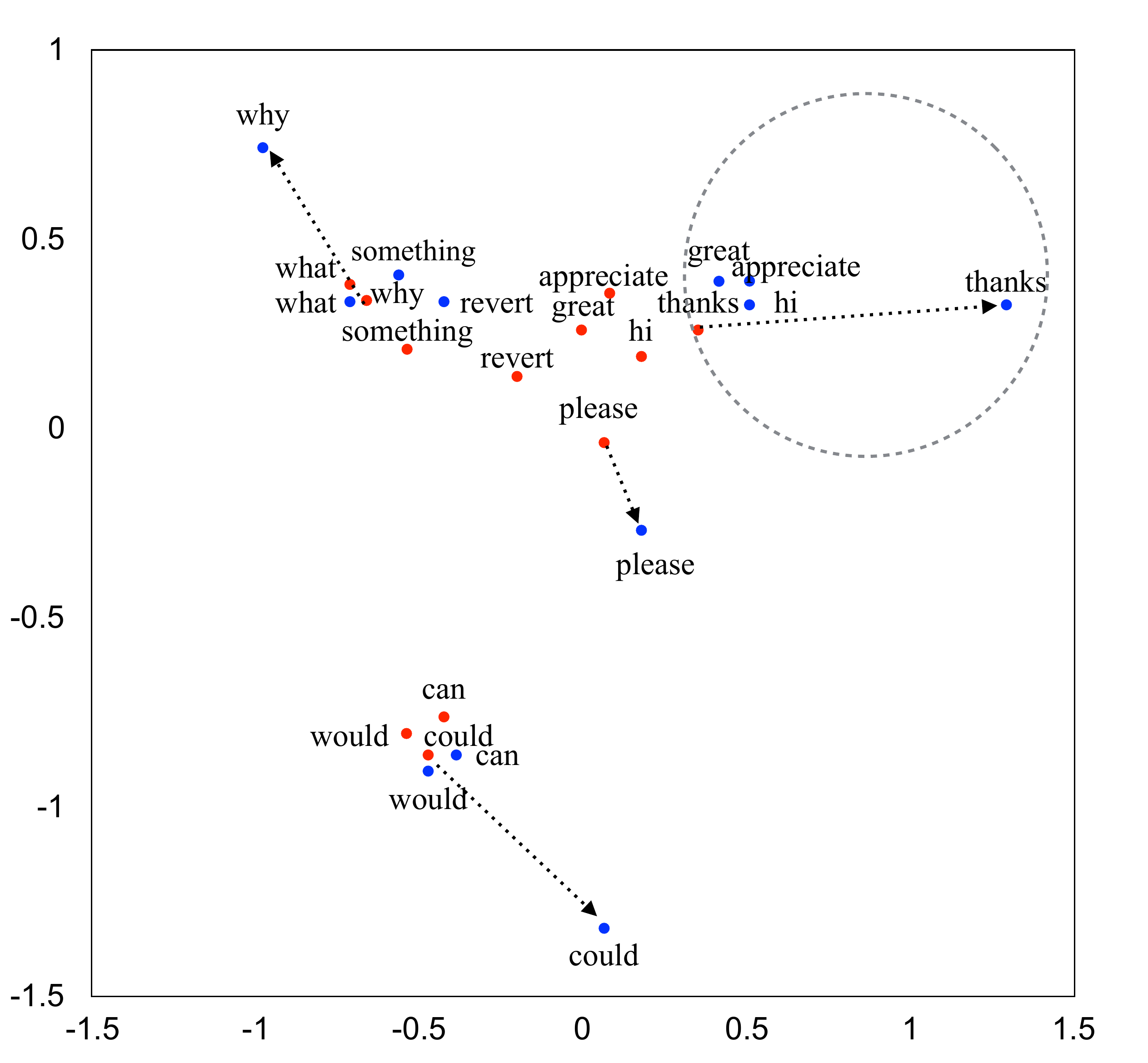}
\centering
\vspace{-10pt}
\caption{Projection before (red) and after (blue) training.}\vspace{-12pt}\label{fig:2dspace}
\end{figure}

We selected key words from~\newcite{danescu2013computational} and from our new activation clusters (~\secref{clusterssection}) and plotted (via PCA) their embedding space positions before and after training, to help us gain insights into specific sentiment transformations.
\figref{fig:2dspace} shows that the most positive keys such as \textit{hi}, \textit{appreciate}, and \textit{great} get clustered even more tightly after training. 
The key \textit{thanks} gets a notably separated position on a positive spectrum, signifying its importance in the NN's decision-making (also depicted via the saliency heatmaps in Sec.~\ref{FDS}).

The indefinite pronoun \textit{something} is located near direct question politeness strategy keys \textit{why} and \textit{what}.
\textit{Please}, as was shown by~\newcite{danescu2013computational}, is not always a positive word because its sentiment depends on its sentence position, and it moves further away from a positive key group.  
Counterfactual Modal keys \textit{could} and \textit{would} as well as \textit{can} of indicative modal get far more separated from positive keys. Moreover, after the training, the distance between \textit{could} and \textit{would} increases but it gets preserved between \textit{can} and \textit{would}, which might suggest that \textit{could} has a far stronger sentiment.  

\subsection{Quantitative Analysis}
\label{sec:quant}
In this section, we present quantitative measures of the importance and polarity of the novelly discovered politeness strategies in the above sections, as well how they explain some of the improved performance of the neural model.

In Table 3 of~\newcite{danescu2013computational}, the pronoun politeness strategy with the highest percentage in top quartile is 2nd Person (30\%). Our extension ~\tabref{tab:extend_table} shows that our novelly discovered Indefinite Pronouns strategy represents a higher percentage (39\%), with a politeness score of -0.13.
Moreover, our Punctuation strategy also turns out to be a top scoring negative politeness strategy and in the top three among all strategies (after Gratitude and Deference). It has a score of -0.71, whereas the second top negative politeness strategy (Direct Start) has a much lower score of -0.43.

Finally, in terms of accuracies, our newly discovered features of Indefinite Pronouns and Punctuation improved the featurized system of~\newcite{danescu2013computational} (see~\tabref{tab:results_table}).\footnote{Our NN visualizations also led to an interesting feature correction. In the 'With Discovered Features' result in~\tabref{tab:results_table}, we also removed the existing pronoun features (\#14-18) based on the observation that those had weaker activation and saliency contributions (and lower top-quartile \%) than the new indefinite pronoun feature. This correction and adding the two new features contributed $\sim$50-50 to the total accuracy improvement.} This reduction of performance gap w.r.t. the CNN partially explains the success of these neural models in automatically learning useful linguistic features. 

%% file: conclusion.tex
\section{Conclusion}
We presented an interpretable neural network approach to  politeness prediction. Our simple CNN model improves over previous work with manually-defined features. More importantly, we then understand the reasons for these improvements via three visualization techniques and discover some novel high-scoring politeness strategies which, in turn, quantitatively explain part of the performance gap between the featurized and neural models.

%% file: acknowledgment.tex
\section*{Acknowledgments}
We would like to thank the anonymous reviewers for their helpful comments. This work was supported by an IBM Faculty Award, a Bloomberg Research Grant, and an NVIDIA GPU donation to MB.

%% file: supplementary.tex

\noindent This is supplementary material for the main paper, where we present more detailed analysis and visualization examples, and our dataset and training details.

\section{Activation Clusters}

\subsection{Rediscovering Existing Strategies}
\paragraph{Gratitude (+)} Respect and appreciation paid to the listener.  
Activation cluster examples: ``\{\textit{thanks for the heads up}"; ``\textit{thank you very much for this kind gesture}"; ``\textit{thanks for help!}"\}\\

\paragraph{Greeting (+)} A welcoming message for the converser. 
Activation cluster examples: \{``\textit{hey, long time no seeing! how's stuff?}"; ``\textit{greetings, sorry to bother you here... }"\} \\

\paragraph{Positive Lexicon (+)} Expressions that build a positive relationship in the conversation and contain positive words from the sentiment lexicon, e.g., \textit{great}, \textit{nice}, \textit{good}.
 Activation cluster examples: \{``\textit{your new map is a great}"; ``\textit{very nice article}"; ``\textit{yes, this is a nice illustration. i 'd love to...}"\} 

\paragraph{Counterfactual Modal (+)} Indirect strategies that imply a burden on the addressee and yet provide a face-saving opportunity of denying the request, usually containing hedges such as \textit{Would it be.../Could you please}. 
Activation cluster examples: \{``\textit{would you be interested in creating an infobox for windmills...?}; ``\textit{would you mind retriveing the bibliographic data?}"\} 

\paragraph{Deference (+)} A way of sharing the burden of a request placed on the addressee.
Activation cluster examples: \{``\textit{nice work so far on your rewrite...}", ``\textit{hey, good work on the new pages...}", ``\textit{good point for the text...}", ``\textit{you make some good points...}"\} 

\paragraph{Direct Question (-)} Questions imposed on the converser in a direct manner with a demand of a factual answer. 
Activation cluster examples: \{``\textit{why would one want to re-create gnaa?}"; ``\textit{what's with the radio , and fist in the air?}"; ``\textit{what level warning is appropriate?}"\} 

\subsection{Extending Existing Strategies}
\paragraph{Counterfactual Modal (+)}
Sentences with \textit{Would you/Could you} get grouped together as expected; but in addition, the cluster contains requests with \textit{Do you mind}. 
Activation cluster examples: ``\{\textit{do you mind having another look?}"; ``\textit{do you mind if i migrate these to your userspace for you?}"\} 

\paragraph{Gratitude (+)}
Our CNN learns a special shade of gratitude, namely it distinguishes a cluster consisting of the bigram \textit{thanks for}.
Activation cluster examples: ``{\textit{thanks for the good advice.}"; ``\textit{thanks for letting me know.}"; ``\textit{fair enough, thanks for assuming good faith}"\} 

\paragraph{Indicative Modal (+)}
The same neuron as for counterfactual modal cluster above also gets activated on gapped 3-grams like \textit{Can you ... please?}, which presumably implies that the combination of a later \textit{please} with future-oriented variants \textit{can/will} in the request gives a similar effect as the conditional-oriented variants \textit{would/could}.  
Activation cluster examples: ``{\textit{can this be reported to london grid, please?}"; ``\textit{can you delete it again, please?}"; ``\textit{good start . can you add more, please?}"\} 

\begin{figure*}[t]
\label{heatmapsupp}
\begin{center}
\setlength{\tabcolsep}{-2pt}
\renewcommand{\arraystretch}{0}
\begin{tabular}{c c c }
\includegraphics[width=5.7cm]{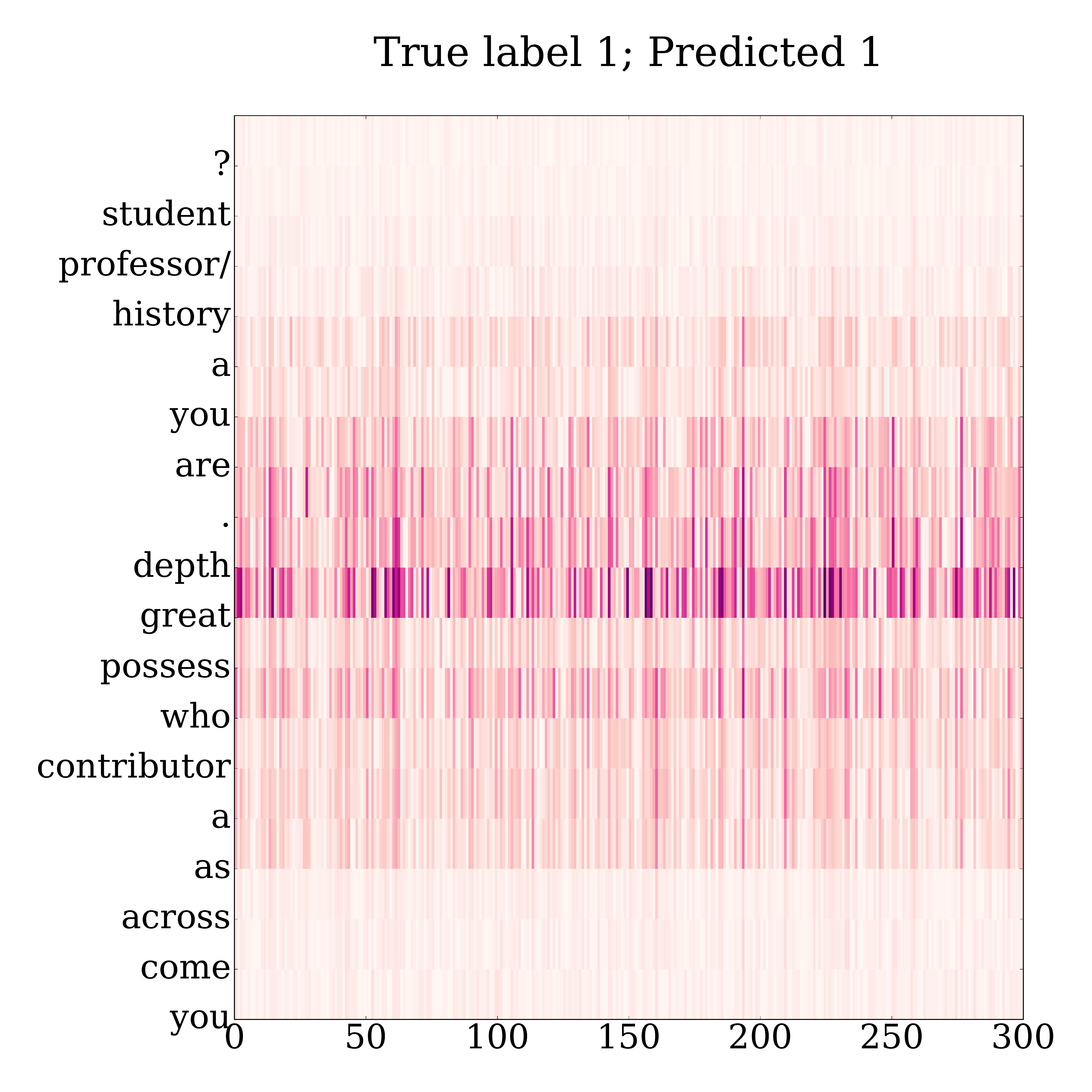} & \includegraphics[width=5.7cm]{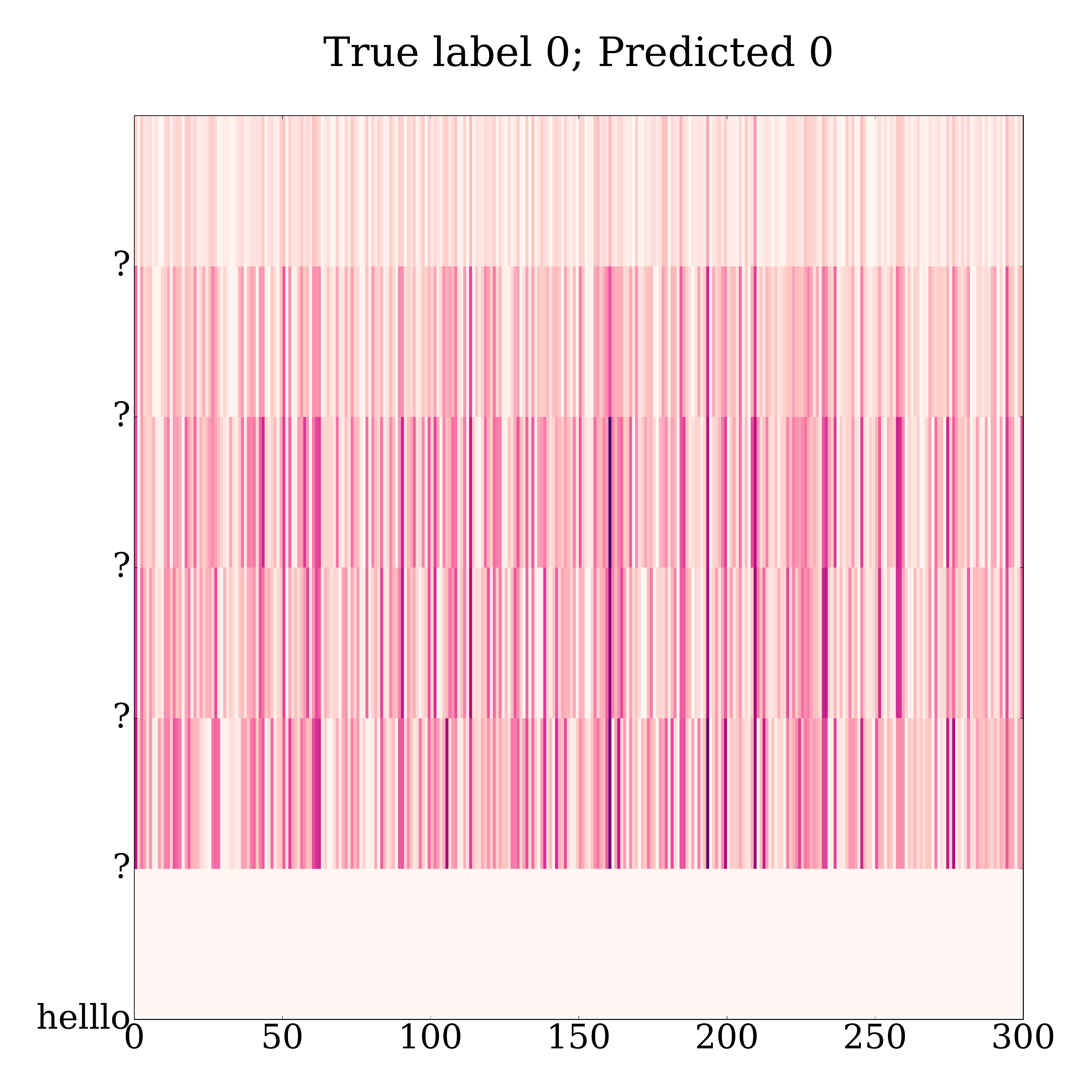} &  \includegraphics[width=5.7cm]{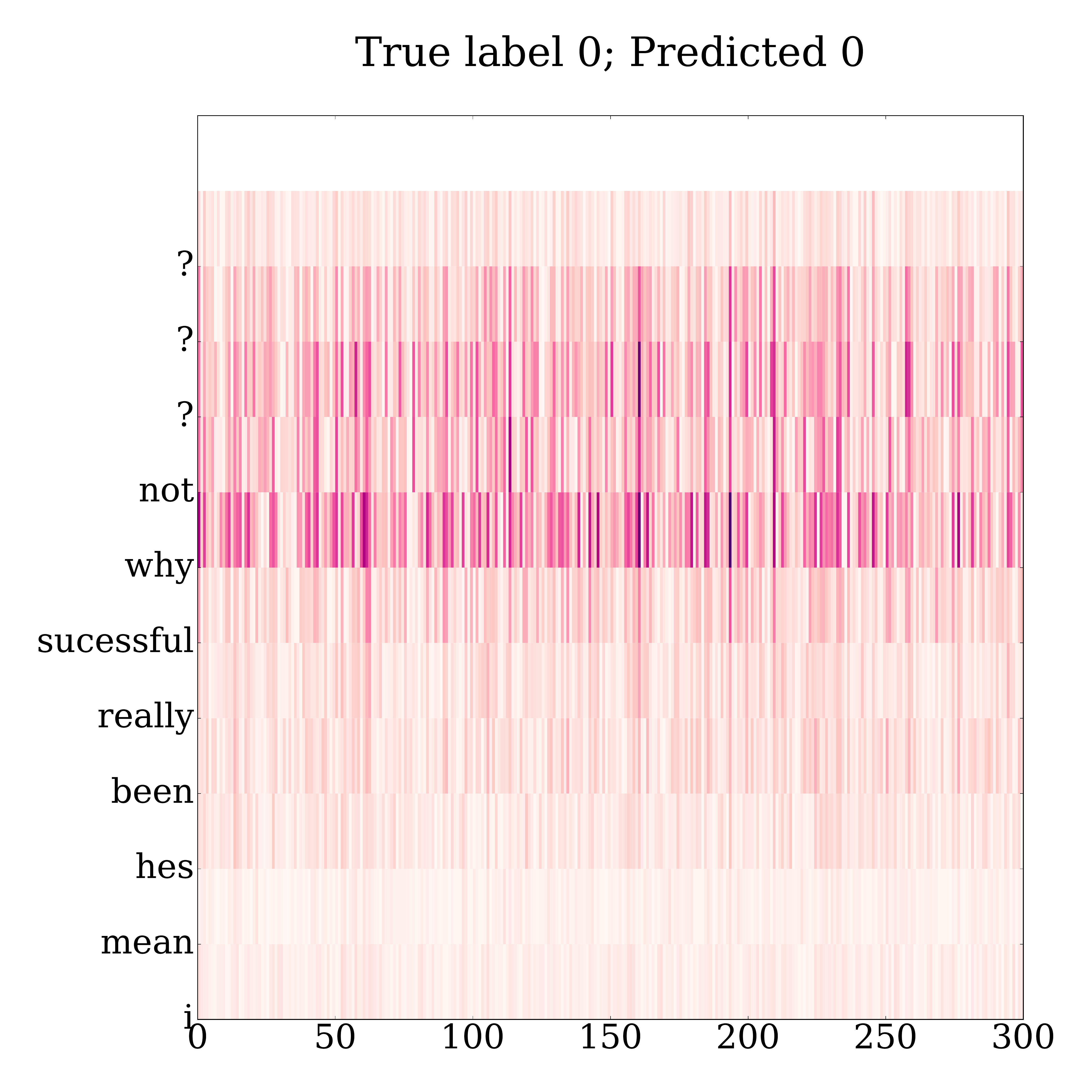} \\
\includegraphics[width=5.7cm]{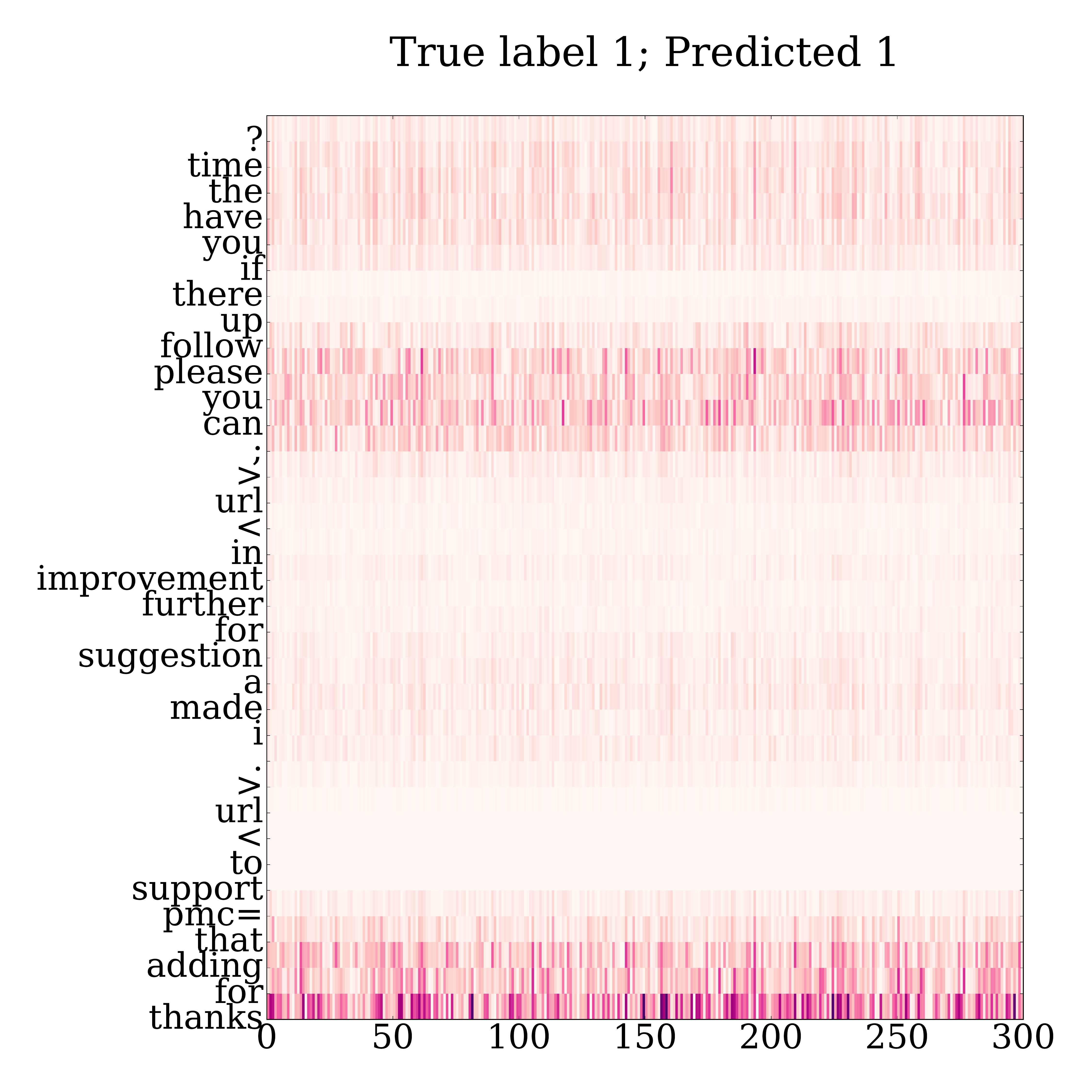} & \includegraphics[width=5.7cm]{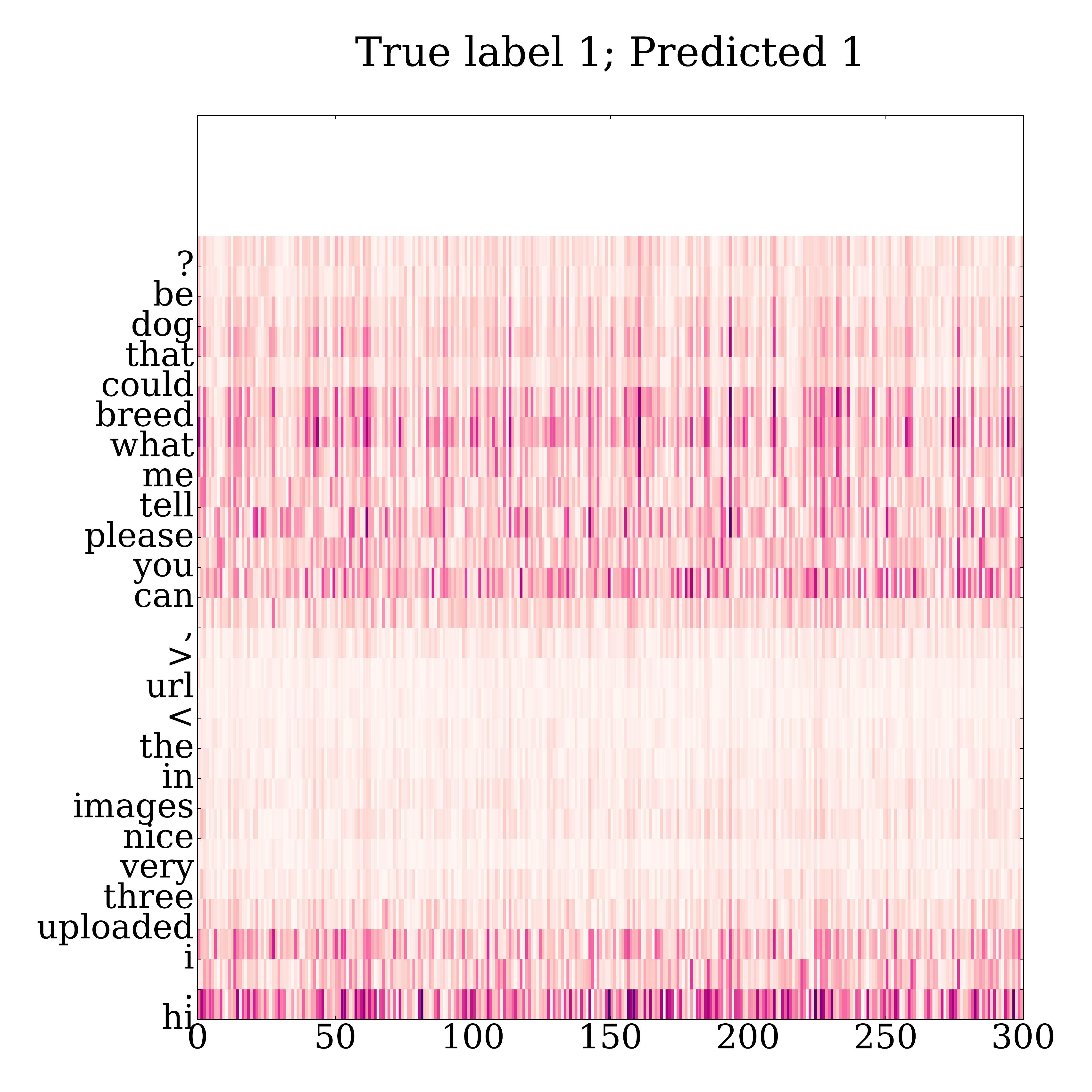} &  \includegraphics[width=5.7cm]{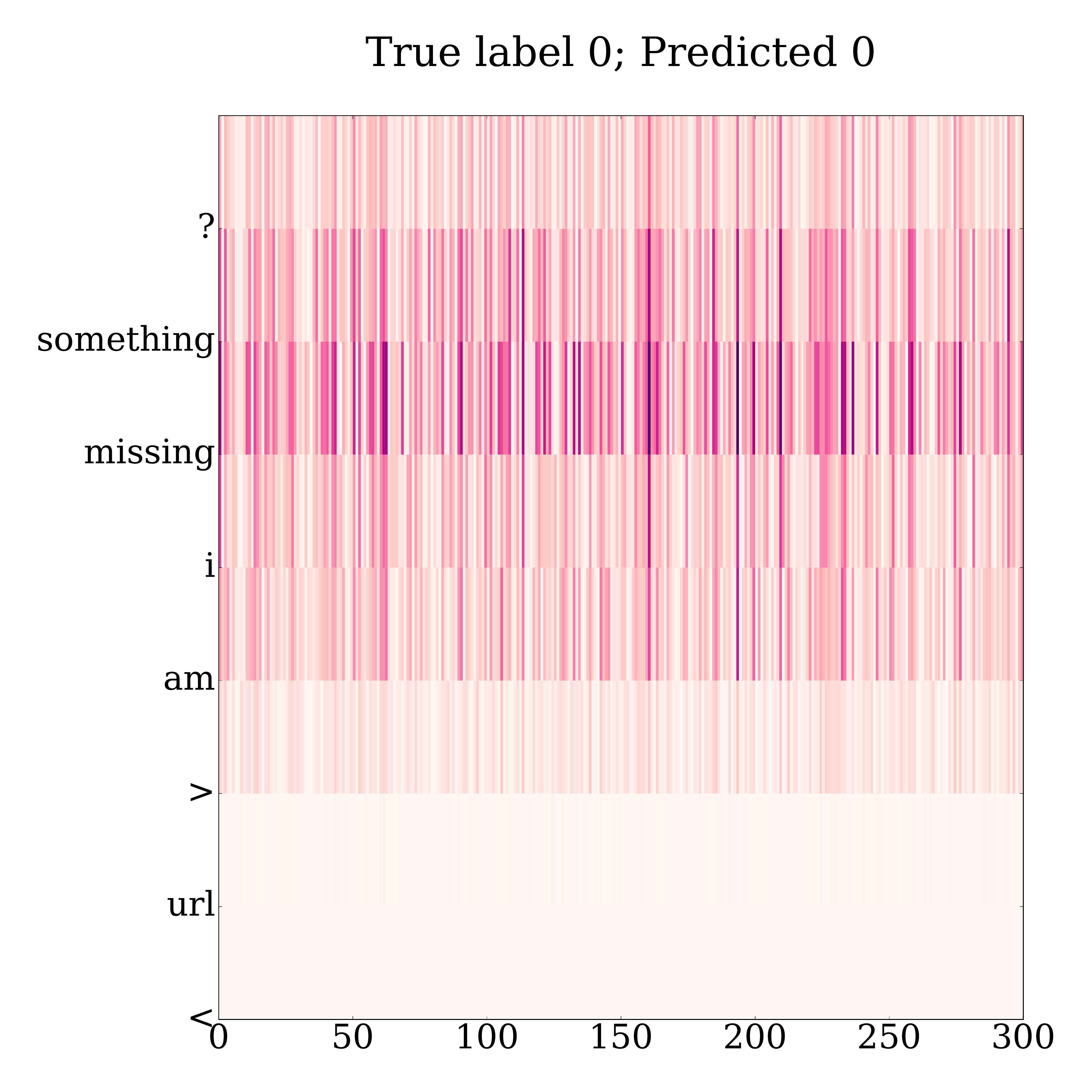} 
\end{tabular}
\end{center}
\vspace{-6pt}
\caption{Additional saliency heatmaps for correctly classified sentences.}
\vspace{-10pt}
\end{figure*}

\subsection{Discovering Novel Strategies}

\paragraph{Indefinite Pronouns (-)} \newcite{danescu2013computational} distinguishes requests with first and second person (plural, starting position, etc.).
However, we find activations that also react to indefinite pronouns such as \textit{something/somebody}.
Activation cluster examples: \{``\textit{am i missing something here?}";  ``\textit{he 's gone. was it something i said?}"; ``\textit{you added the tag and then mentioned it on talk- you did not gain consensus first or even wait for anyone to discuss it. }"; ``\textit{but how can something be both correct and incorrect}" \} 

\paragraph{Punctuation (-)}  Though non-characteristic in direct speech, punctuation appears to be an important special marker in online communities, which in some sense captures verbal emotion in text.  
One of our neuron clusters gets activated on question marks ``\textbf{???}" and one on ellipsis ``\textbf{...}".
Activation cluster examples of question marks:  \{``\textit{now???}"; ``\textit{original article????}"; ``\textit{helllo?????}"\}
Activation cluster examples of ellipsis: \{``\textit{ummm , it 's a soft redirect. a placeholder for a future page \textbf{...} is there a problem ?}"; ``\textit{Indeed \textbf{...} the trolls just keep coming.}"; ``\textit{I can't remember if i asked/checked to see if it got to you? so \textbf{...} did it ?}"\}


\section{First Derivative Saliency}
In Fig.~1, we show some additional examples of saliency heatmaps.
In the first heatmap, we see a clear example of the Positive Lexicon politeness strategy. The key \textit{great} captures most of the weight for the final decision making. Note that, in particular, the question mark in this case provides no influence. 
Contrast that to the second figure, which echos back the proposed negative politeness strategy on punctuation from Section 6.1.3. Initial question marks give a high influence in magnitude for the negative predicted label. 
In the third example, we see that these punctuation markers still provide a lot of emphasis. For instance, other words such as \textit{really}, \textit{successful} and a personal pronoun \textit{I} have very little impact. Overall, this exemplifies Direct Question strategy since most of the focus is on \textit{why}. 

As was noted in the embedding space transformations discussion, the Gratitude key \textit{thanks} with a preposition \textit{for} has a much stronger polarity than other positive politeness keys in the fourth heatmap. Indeed, \textit{can you please} does not nearly provide as much value. 
In the fifth heatmap, the sensitivity of the final score comes more from the greeting, namely \textit{hi}, as compared to the phrase \textit{can you please tell me} or positive lexicon \textit{very nice}. These results match the politeness score results in Table 3 of~\newcite{danescu2013computational}, where the Greeting strategy has a score of 0.87 compared to 0.49 for Please strategy and 0.12 for Positive Lexicon strategy.
The sixth and last heatmap demonstrates the contribution of indefinite pronouns. In this case phrase \textit{am I missing something} with the focus on the latter two words decides the final label prediction. 

\section{Dataset and Training Details}
We split the Wikipedia and Stack Exchange datasets of~\newcite{danescu2013computational} into training, validation and test sets with 70\%, 10\%, and 20\% of the data respectively (after random shuffling). Therefore, the final split for Wikipedia is 1523, 218 and 436; and for Stack Exchange it is  2298, 328, and 657, respectively. We will make the dataset split indices publicly available.

We use 300-dim pre-trained {\small\tt word2vec} embeddings as input to the CNN~\newcite{mikolov2014word2vec}, and then allow fine-tuning of the embeddings during training. 
All sentence tokenization is done using NLTK~\cite{bird2006nltk}. For words not present in the pre-trained set, we use uniform unit scaling initialization. 

We implement our model using a python version of TensorFlow~\cite{tensorflow2015-whitepaper}. Hyperparameters, e.g., the mini-batch size, learning rate, optimizer type, and dropout rate were tuned using the validation set of Wikipedia via grid search.\footnote{Grid search was performed over dropout rates from 0.1 to 0.9 in increments of 0.1; four learning rates from 1e-1 to 1e-4; Adam, SGD, and AdaGrad optimizers; filter windows sizes from 1 to 3-grams; and features maps ranging from 10 to 200 with an incremental step of 20.} The final chosen values were a mini-batch size of 32, a learning rate of 0.001 for the Adam Optimizer, a dropout rate of 0.5, filter windows are 3, 4, and 5 with 75 feature maps each, and ReLU as non-linear transformation function~\cite{nair2010rectified}. For convolution layers, we use valid padding and strides of all ones.